\pdfoutput=1
\documentclass[11pt]{article}
\usepackage[]{EACL2023}
\usepackage{times}
\usepackage{graphicx}
\usepackage{latexsym}
\usepackage[T1]{fontenc}
\usepackage[utf8]{inputenc}
\usepackage{microtype}
\usepackage{hyperref}
\usepackage{inconsolata}
\usepackage{ctable}

\title{Two-stage Pipeline for Multilingual Dialect Detection}

\author{Ankit Vaidya \and
    Aditya Kane \\
  Pune Institute of Computer Technology, Pune \\
  \texttt{\{ankitvaidya1905, adityakane1\}@gmail.com} }

\begin{document}
\maketitle
\begin{abstract}
% Ankit
Dialect Identification is a crucial task for localizing various Large Language Models. This paper outlines our approach to the VarDial 2023 DSL-TL shared task. Here we have to identify three or two dialects from three languages each which results in a 9-way classification for Track-1 and 6-way classification for Track-2 respectively. Our proposed approach consists of a two-stage system and outperforms other participants' systems and previous works in this domain. We achieve a score of 58.54\% for Track-1 and 85.61\% for Track-2. Our codebase is available publicly\footnote{\url{https://github.com/ankit-vaidya19/EACL_VarDial2023}}. 
\end{abstract}

\section{Introduction}
% Aditya
Language has been the primary mode of communication for humans since the pre-historic ages. Studies have explored the evolution of language and outlined mathematical models that govern the intricacies of natural language \cite{evolution_of_language, mystery_of_language_evolution}. Inevitably, as humans established civilization in various parts of the world, this language was modified by, and for the group of people occupied by that particular geographical region. This gave rise to multiple national dialects of the same language. 

The VarDial workshop \cite{2023-findings-vardial} (co-located with EACL 2023) explores various dialects and variations of the same language. We participated in the Discriminating Between Similar Languages - True Labels (DSL-TL) shared task. In this task, the participants were provided with data from three languages, with each language having three varieties. This shared task consisted of two tracks -- Track-1 featuring nine-way classification and Track-2 featuring six-way classification. The second track included two particular national dialects of each language (eg. American English and British English), and the first track had one general label (English) in addition to the aforementioned two. 

We ranked $1^{st}$ in both of the tracks. Moreover, we beat the next best submission by a margin of 4.5\% in the first task and 5.6\% in the second task.We were the only team to surpass the organizer baseline scores. We present our winning solution in this paper. We used an end-to-end deep learning pipeline which consisted of a language identification model and three language-specific models, one for each language. We converged upon the best combination by doing an elaborate analysis of various models available. Furthermore, in this work we also analyze the performance of the pipeline as a whole and also provide an ablation study. Lastly, we provide some future directions in this area of research.

\section{Related Work}
\begin{figure}
    \centering
    \includegraphics[width=\linewidth]{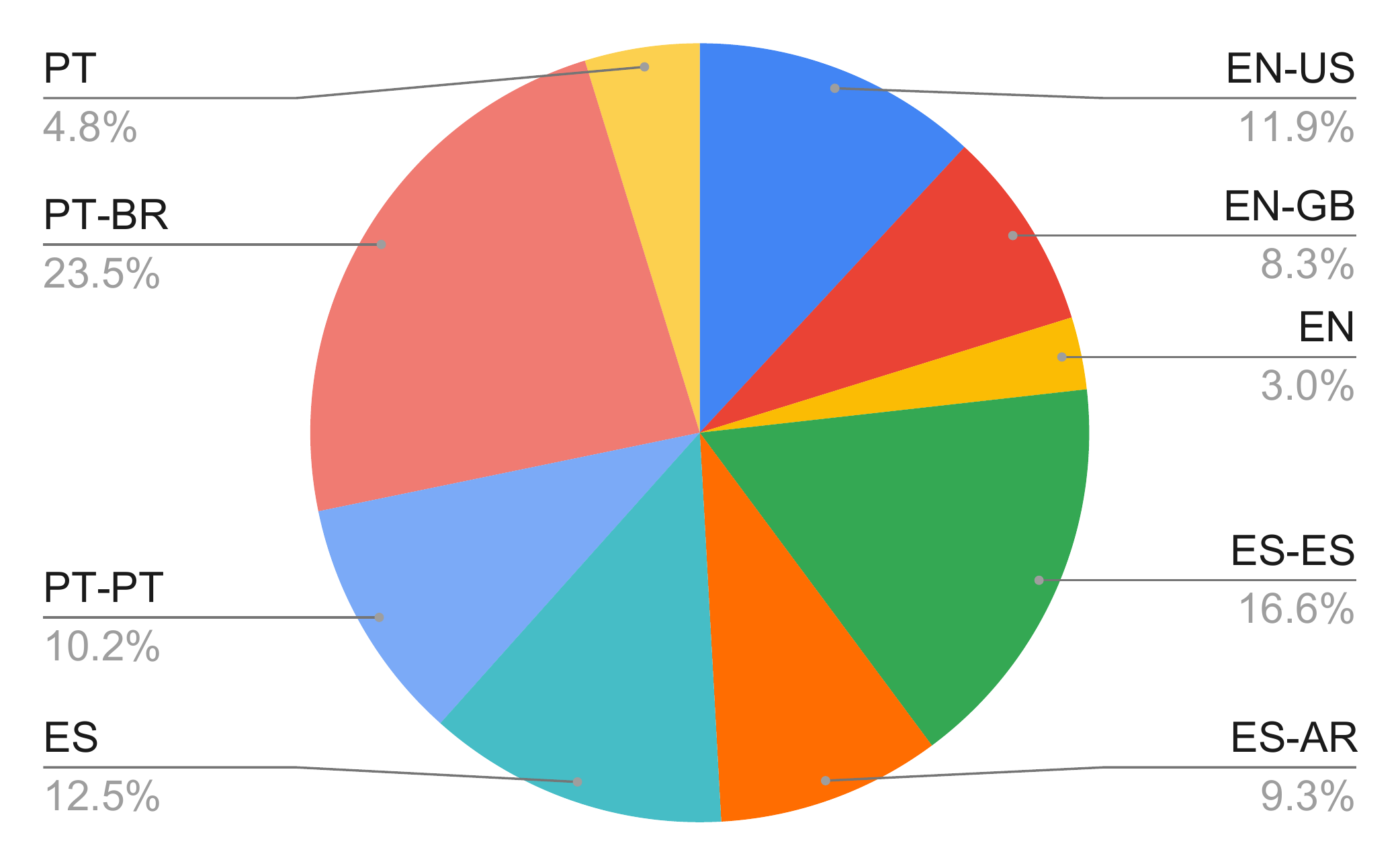}
    \caption{Class distribution of dialects}
    \label{fig:data_distribution}
\end{figure}

\begin{figure*}[!tbh]
    \centering
    \includegraphics{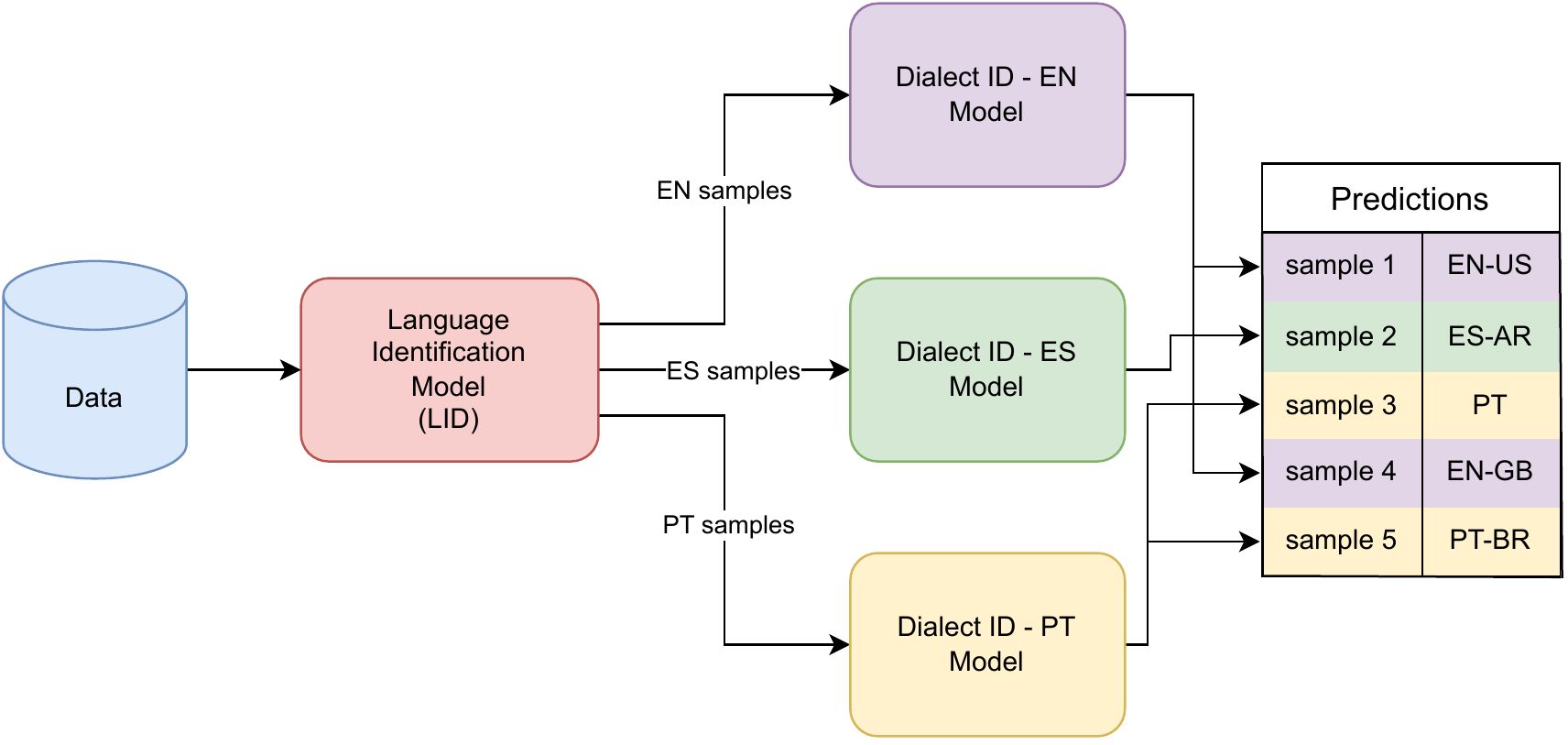}
    \caption{System diagram for dialect classification.The LID classifies the input into one of 3 languages.The sample is then further classified into dialects by language specific models.}
    \label{fig:system_description}
\end{figure*}
% Aditya
The present literature encompasses various aspects of dialect identification. We study this from three perspectives: large language models, language identification and dialect classification problems. 

\subsection{Large Language Models}

The success of transformers and BERT \cite{devlin-etal-2019-bert} based models was inevitable since the initial boom of the transformer \cite{attention} model. In recent years, many other architectures like RoBERTa \cite{Liu2019RoBERTaAR} and ELECTRA \cite{electra} have further pushed the state-of-the-art in this domain. Moreover, autoregressive models like GPT \cite{Radford2018ImprovingLU} and GPT-2 \cite{Radford2019LanguageMA} have also shown their prowess. Multilingual versions of RoBERTA, namely XLM-RoBERTa \cite{conneau-etal-2020-unsupervised} are also available. Lastly, language specific models like Spanish BERT \cite{BERTIN} and Portuguese BERT \cite{souza2020bertimbau} are available as well. Our winning solution makes use of these large language models trained on specific languages.

\subsection{Language Identification Models}

Many multilingual language identification models have been developed in order to classify the language of the input sentence beforehand. Even though the initial works used $n$-gram models and generative mixture models \cite{lui-etal-2014-automatic, baldwin-lui-2010-language, al-badrashiny-diab-2016-lili, bernier-colborne-etal-2021-n, bestgen-2021-optimizing} or even conditional random fields \cite{al-badrashiny-diab-2016-lili} and other classical machine learning methods like naive bayes \cite{jauhiainen-etal-2020-uralic, jauhiainen-etal-2022-italian, jauhiainen-etal-2021-naive}, modern methods have shifted to the use of deep learning for language identification \cite{lide, lid_malyalam, romero21_iberspeech, bernier-colborne-etal-2022-transfer}. Recent works have mainly focused on deep learning based language identification, where handling codemixed data is a big challenge in the domain. For our experiments, we use a version of XLM-RoBERTa finetuned on a language identification dataset\footnote{This model is available \href{https://huggingface.co/papluca/xlm-roberta-base-language-detection}{here} and dataset is available \href{https://huggingface.co/datasets/papluca/language-identification}{here}}. This model has near-perfect test accuracy of 99.6\%. 

\subsection{Dialect Classification}

Dialect classification has been previously solved using statistical methods like Gaussian Mixture Models and Frame Selection Decoding or Support Vector Machines (SVM) \cite{arabic_dialect, tillmann-etal-2014-improved}. It has been explored relatively sparsely, mostly in the case for local languages \cite{kannda_dialect}. Deep learning approaches have been explored in previous editions of the VarDial workshop shared tasks \cite{rebeja-cristea-2020-dual} and otherwise \cite{transformer_based_arabic_dialect}. Dialect classification was also explored previously as a part of other shared tasks \cite{khered-etal-2022-building}. We want to stress that given the multilingual nature of the dataset, using the present methods directly was not an option. In our work, although we take inspiration from the previous works, we propose a novel system that surpasses the performance of the previous systems by a large margin.

%============================== Final 3 way ================================%
\begin{table*}[h!]
\begin{tabular}{ccccc}
 \toprule
\textbf{Model-EN} & \textbf{Model-ES}            & \textbf{Model-PT}            & \textbf{Validation F1} & \textbf{Test F1} \\ \midrule
RoBERTa$_{base}$       & Spanish RoBERTa$_{base}$          & XLM-RoBERTa$_{base}$              & 60.74\%                  &        -            \\
RoBERTa$_{base}$       & Spanish BERT$_{base}$             & XLM-RoBERTa$_{base}$              & 60.05\%                  &        -           \\
BERT$_{base}$          & Spanish RoBERTa$_{base}$          & XLM-RoBERTa$_{base}$              & 59.08\%                  &        -           \\
BERT$_{base}$          & Spanish BERT$_{base}$             & Portuguese BERT$_{base}$          & 60.17\%                  &        -           \\
BERT$_{base}$          & Spanish BERT$_{base}$             & XLM-RoBERTa$_{base}$              & 58.40\%                  &        -           \\\midrule
\textbf{BERT$_{base}$} & \textbf{Spanish RoBERTa$_{base}$} & \textbf{Portuguese BERT$_{base}$} & \textbf{60.85\%}         & \textbf{58.54\%}   \\
RoBERTa$_{base}$       & Spanish RoBERTa$_{base}$          & Portuguese BERT$_{base}$          & 62.51\%                  & 58.09\%            \\
RoBERTa$_{base}$       & Spanish BERT$_{base}$             & Portuguese BERT$_{base}$          & 61.83\%                  & 57.03\%      \\     \bottomrule
\end{tabular}
\caption{Our complete results for Track-1 using the two-stage dialect detection pipeline. Model-* denotes the language of the models used for the experiments.}
\label{tab:final_3_way}
\end{table*}

%============================== Final 2 way ================================%
\begin{table*}[!ht]
\begin{tabular}{ccccc}
 \toprule
\textbf{Model-EN}    & \textbf{Model-ES}            & \textbf{Model-PT}            & \textbf{Validation F1} & \textbf{Test F1} \\ \midrule
RoBERTa$_{base}$          & XLM-RoBERTa$_{base}$              & Portuguese BERT$_{base}$          & 80.84\%                  & -                  \\  
RoBERTa$_{base}$          & XLM-RoBERTa$_{base}$              & XLM-RoBERTa$_{base}$              & 79.16\%                  & -                  \\  
BERT$_{base}$             & Spanish RoBERTa$_{base}$          & XLM-RoBERTa$_{base}$              & 82.22\%                  & -                  \\  
BERT$_{base}$             & XLM-RoBERTa$_{base}$              & Portuguese BERT$_{base}$          & 80.55\%                  & -                  \\  
BERT$_{base}$             & XLM-RoBERTa$_{base}$              & XLM-RoBERTa$_{base}$              & 78.87\%                  & -                  \\ \hline
\textbf{RoBERTa$_{base}$} & \textbf{Spanish RoBERTa$_{base}$} & \textbf{Portuguese BERT$_{base}$} & \textbf{84.19\%}         & \textbf{85.61\%}   \\  
BERT$_{base}$             & Spanish RoBERTa$_{base}$          & Portuguese BERT$_{base}$          & 83.90\%                  & 85.11\%            \\  
RoBERTa$_{base}$          & Spanish RoBERTa$_{base}$          & XLM-RoBERTa$_{base}$              & 82.51\%                  & 83.68\%            \\  \bottomrule
\end{tabular}
\caption{Our complete results for Track-2 using the two-stage dialect detection pipeline. Model-* denotes the language of the models used for the experiments.}
\label{tab:final_2_way}
\end{table*}

\section{Data}
% Ankit
% Data distribution: Pie chart / bar graph
% # data points, train test split, data imbalance
% optional: Average length of a sample for every class, total number of tokens in the dataset
% \begin{multline}

The dataset \cite{zampieri2023language} contained a total of 11,610 sentences belonging to 3 languages: English (EN), Spanish (ES), Portuguese (PT) and each language had 3 corresponding varieties. The varieties for English were: American English ($EN-US$), British English ($EN-GB$) and Common English Instances ($EN$). Similarly varieties corresponding to Spanish and Portuguese were: European/Peninsular Spanish ($ES-ES$), Argentine Spanish ($ES-AR$), Common Spanish Instances ($ES$) and European Portuguese ($PT-PT$), Brazilian Portuguese ($PT-BR$), Common Portuguese Instances ($PT$). These were divided into a training set containing 8,745 sentences and the validation set containing 2,865 sentences. The system was evaluated on a separate testing set containing 1,290 sentences. 
This dataset has acute class imbalance. We observed that the class PT-BR had the most number of samples (2,724) and the class EN had the least number of samples (349), and thus the imbalance ratio was almost 1:8. We have illustrated the data distribution in Figure \ref{fig:data_distribution}. We tried to mitigate this imbalance using over-sampling and weighted sampling methods. However, the improved data sampling method did not affect the performance.

\section{System Description}

% Ankit
% Diagram: Overall system diagram
% Explain the diagram
% LID cha input output and how do you parse it
% Individual models setup, weight balancing
% \begin{multline}
\label{sec:sys_desc}

This was a problem of multi-class classification having 9 classes for Track-1 and 6 classes for Track-2. The samples were belonging to 3 languages having 3 varieties each, so the classification pipeline was made in 2 stages. The Language Identification (LID) model which is the first stage classifies the sentence into 3 languages: English (EN), Spanish (ES) and Portuguese (PT). The LID is a pretrained XLM-RoBERTa that is fine-tuned for the task of language identification. It is able to classify the input sentence into 20 languages. We classify and separate the samples according to their language. The samples corresponding to the specific languages are then fed into the language specific models for dialect identification. For dialect identification we have used models like BERT and RoBERTa with a linear layer connected to the pooler output of the models. Then fine-tuning is done on the models for dialect identification using the samples corresponding to the specific languages. For the task of dialect identification we experimented with several pretrained models like XLM-RoBERTa, BERT, ELECTRA, GPT-2 and RoBERTa. All models were fine-tuned for 20 epochs with a learning rate of 1e-6 and weight decay 1e-6 with a batch size of 8. The best performing model checkpoint was chosen according to the epoch-wise validation macro-F1 score.

\section{Experiments and Results}

%============================== Individual 3 way ================================%
\setlength{\tabcolsep}{3pt}
\begin{table}[]
\begin{tabular}{cccc}\toprule
\textbf{Lg} & \textbf{Model}      & \textbf{Train F1} & \textbf{Val F1}   \\ \midrule
\textbf{EN}   & \textbf{RoBERTa}      & \textbf{79.74\%}      & \textbf{71.34\%}          \\
EN                & BERT              & 80.71\%                & 70.19\%   \\ 
EN                & ELECTRA            & 65.02\%                & 66.60\%    \\ 
EN                & XLM-RoBERTa        & 71.64\%                & 66.12\%     \\
EN                & GPT-2                & 56.78\%                & 49.74\%      \\
\midrule
\textbf{ES}      & \textbf{Spanish RoBERTa}    & \textbf{74.36\%}    &\textbf{62.96\%} \\
ES                & XLM-RoBERTa        & 59.46\%          & 61.58\%    \\
ES                & Spanish BERT       & 67.40\%          & 60.76\%    \\
ES                & Spanish GPT-2        & 34.33\%           & 46.11\%\\
\midrule
\textbf{PT}      & \textbf{Portuguese BERT}    & \textbf{67.63\%}       & \textbf{55.15\%}   \\
PT                & XLM-RoBERTa        & 64.33\%                                 & 48.46\%    \\
PT                & Portuguese ELECTRA & 62.11\% & 46.34\%    \\
PT                & Portuguese GPT-2     & 38.52\%                             & 34.19\%  \\
\bottomrule
\end{tabular}
\caption{Performance on Track-1 validation dataset of individual models used in the two-stage pipeline. "Lg" stands for language of the model used.}
\label{tab:individual_3_way}

\end{table}

%============================== Individual 2 way ================================%
\begin{table}[]
\begin{tabular}{cccc}\toprule
\textbf{Lg} & \textbf{Model}               & \textbf{Train F1} & \textbf{Val F1} \\ \midrule
\textbf{EN}       & \textbf{RoBERTa}         & \textbf{91.70\%}    & \textbf{88.75\%}         \\
EN                & BERT                     & 94.24\%             & 88.32\%                  \\
EN      & XLM-RoBERTa     & 87.61\%    & 84.68\%         \\
\midrule
\textbf{ES}       & \textbf{Spanish RoBERTa} & \textbf{96.05\%}    & \textbf{87.05\%}         \\
ES                & XLM-RoBERTa              & 89.25\%             & 80.29\%                  \\
\midrule
\textbf{PT}       & \textbf{Portuguese BERT} & \textbf{89.49\%}    & \textbf{79.21\%}         \\
PT                & XLM-RoBERTa              & 81.61\%             & 75.91\%                \\
\bottomrule
\end{tabular}
\caption{Performance on Track-2 validation dataset of individual models used in the two-stage pipeline. "Lg" stands for language of the model used.}
\label{tab:individual_2_way}
\end{table}

\begin{table}[]
\begin{tabular}{cccc}\toprule
\textbf{Lg} & \textbf{Model}      & {\textbf{Adapted F1}} & \textbf{F.T. F1} \\ \midrule
EN                & RoBERTa         & 85.20\%                                            & 88.75\%                                      \\
EN                & BERT            & 83.21\%                                            & 88.32\%                                      \\
EN                & XLM-RoBERTa     & 81.21\%                                            & 84.68\%                                      \\
\midrule
ES                & Spanish RoBERTa & 78.45\%                                            & 87.05\%                                      \\
ES                & XLM-RoBERTa     & 66.89\%                                            & 80.29\%                                      \\
\midrule
PT                & Portuguese BERT & 72.17\%                                            & 79.21\%                                      \\
PT                & XLM-RoBERTa    & 71.89\%                                            & 75.91\% \\
\bottomrule
\end{tabular}
\caption{Comparative results of two-way classification using the finetuned (F.T.) predictions and predictions adapted from three-way classification models.}
\label{tab:adapted_2_way}
\end{table}
\setlength{\tabcolsep}{6pt}

\subsection{Experiments using Large Language Models}
For the task of Dialect Identification we have tried various language specific models like XLM-RoBERTa, BERT, ELECTRA, RoBERTa and GPT-2. The base variant of all these models were used and all the models were used through the HuggingFace \cite{wolf-etal-2020-transformers} library. The pooler output of these models was passed through a linear layer and the models were fine-tuned. First, we experimented with different models for Track-1. All the models were trained for 20 epochs with learning rate 1e-6, weight decay 1e-6 and a batch size of 8. We used XLM-RoBERTa as the baseline for all 3 languages. The best performing models for the English language were RoBERTa and BERT whereas GPT-2 was the worst performing. Similarly the language specific versions of RoBERTa and BERT performed well for the Spanish and Portuguese respectively. Overall the worst performing model was GPT-2 across all 3 languages. The validation F1 scores are present in Table \ref{tab:individual_3_way}. The two best-performing models for every language were chosen for Track-2. The same procedure as specified above was used and the F1 scores are present in Table \ref{tab:individual_2_way}. The train and validation F1 scores for 2-class classification are higher for all models as compared to the F1 score of the same models for 3-class classification. This was mainly due to the poor representation and accuracy of classification of the third class. We observed symptoms of overfitting in all models after 12-15 epochs and the best validation F1 score was obtained in the range of 4-8 epochs.

\subsection{LID experiments}
\label{subsec:LID_experiments}
The pipeline for dialect identification is divided into two parts as the sentences in the dataset belong to different languages. The stages are described in Section \ref{sec:sys_desc}. The XLM-RoBERTa we have used for language classification has a test accuracy of 99.6\% meaning it correctly classifies all input sentences and hence, can be considered as a perfect classifier. For the final pipeline we experimented using the two best performing models for each language in Track-1 and Track-2. For both the tracks we experimented with all 8 ($2^3$) possible combinations of models and calculated the validation F1 score for the combined validation dataset which had sentences belonging to all languages. The validation scores for Track-1 and Track-2 are shown in Table \ref{tab:final_3_way} and Table \ref{tab:final_2_way} respectively. For both the tracks, the three pipelines with the best validation F1 scores were chosen for submission.

\subsection{Using 3-way classifier as a 2-way classifier}

In Track-1, participants are expected to train a classifier which classifies amongst 9 classes, and in Track-2, participants are expected to train a classifier which classifies amongst 6 classes. These 6 classes are a proper subset of the 9 classes from Track-1. Thus, an intuitive baseline for Track-2 is to use the model finetuned for Track-1, whilst considering only the relevant classes for the latter task. The classes $EN$, $ES$ and $PT$, i.e. the classes without any national dialect associated with them are not included in Track-2 as compared to Track-1. Thus, we calculate the predictions for the Track-2 validation dataset using the models for Track-1 and exclude the metrics for Track-1 specific classes to get the metrics for this "adapted" 2-way classification. We show the results of this experiment in Table \ref{tab:adapted_2_way} and observe that, as expected, the adapted 2-way classification performs worse compared to the explicitly finetuned variant.

\subsection{Results for Track-1 and Track-2}

We now present our experiments and their performance for both tracks. Our experiments for Track-1 are described in Table \ref{tab:final_3_way} and our experiments for Track-2 are described in Table \ref{tab:final_2_way}. The participants were allowed three submissions for evaluation on the test set, so we submitted predictions using the three systems which performed the best on the validation set. As mentioned in Section \ref{subsec:LID_experiments}, we performed $2^3$, i.e. a total of 8 experiments using the two best models for each language. We observed that RoBERTa$_{base}$ on English, Spanish BERT$_{base}$ on Spanish and Portuguese BERT$_{base}$ performed the best on the testing set for Track-1. The same combination, with RoBERTa$_{base}$ for English, worked best for Track-2. All of our submissions were the top submissions for each track, which surpassed the next best competitors by a margin of 4.5\% and 5.6\% for Track-1 and Track-2 respectively. 

\subsection{Ablation of best submissions}

\begin{figure}
    \centering
    \includegraphics[width=\linewidth]{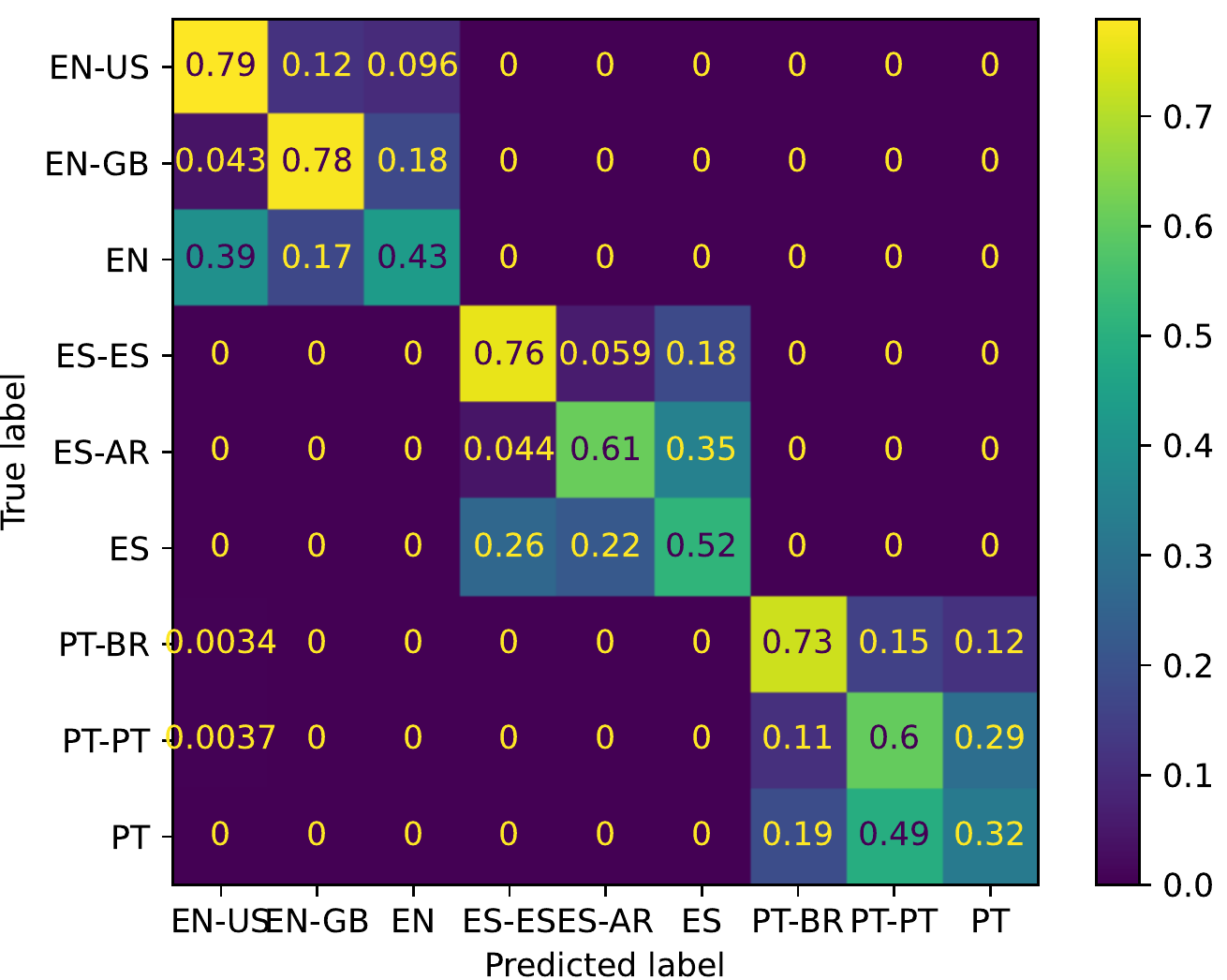}
    \caption{Confusion matrix of 9-way classification. Note that rows are normalized according to the number of samples is that class.}
    \label{fig:confmat_3way}
\end{figure}

\begin{figure}
    \centering
    \includegraphics[width=\linewidth]{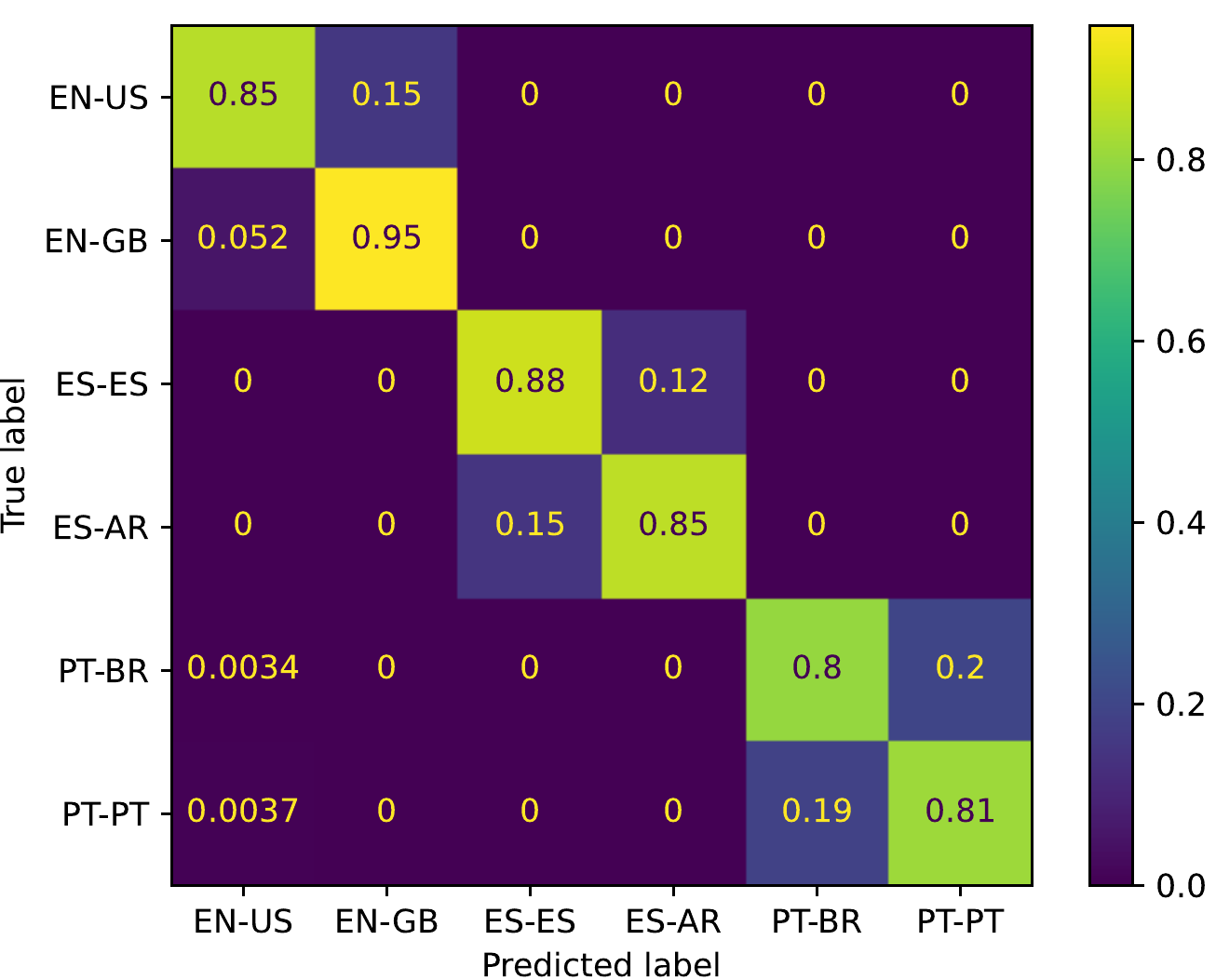}
    \caption{Confusion matrix of 6-way classification. Note that rows are normalized according to the number of samples is that class.}
    \label{fig:confmat_2way}
\end{figure}

We hereby make some observations of our submissions and other experiments. To assist this, we plot the confusion matrices of our best submissions for Track-1 and Track-2 in Figures \ref{fig:confmat_3way} and \ref{fig:confmat_2way} respectively. Note that these confusion matrices have their rows (i.e. true labels axes) normalized according to the number of samples in the class.  Here are observations from our experiments:

\begin{enumerate}
    \item \textbf{BERT-based models outperform other models across all languages:} We observe that BERT-based models outperform ELECTRA-based and GPT-2-based models, as shown in Table \ref{tab:individual_3_way}. We speculate this is because of the inherent architecture of BERT, which combines semantic learning with knowledge retention. This combination of traits is particularly useful for this task.
    \item \textbf{Common labels perform the worst across all languages:} We observe that the common labels $EN$, $ES$ and $PT$ perform the worst, both in the individual as well as the two-stage setup. We hypothesize this is because of the absence of dialect specific words, or words that are specific to the geographical origin of the national dialect (for example, "Yankees" for EN-US and "Oxford" for EN-GB).
    
    \item \textbf{English models work better than models of other languages:} It can be noted from Figures \ref{fig:confmat_2way} and \ref{fig:confmat_3way} that the English models have the best performance across all classes. This can be attributed to two reasons: absence of national dialect specific words and lesser pretraining data in the case of Portuguese.

    \item \textbf{British English is most correctly classified class:} We can observe that the Spanish or Portuguese models make equal number of mistakes in the case of either national dialect, in the case of Track-2 (see Figure \ref{fig:confmat_2way}). However, in the case of English, the label $EN-GB$ is correctly classified for more than 95\% of the cases. We speculate this is because British English involves slightly distinctive grammar and semantics, which help the model separate it from other classes. 
    
    \item \textbf{The proposed 2-step method is scalable for multiple language dialect classification:} We can strongly assert that the novel 2-step deep learning method for multilingual dialect classification is a scalable method for the task due to two specific reasons: firstly, the multilingual models (like XLM-RoBERTa) might not have the vocabulary as well as the learning capabilities to learn the minute differences between individual dialects. Secondly, this system can be quickly expanded for a new language by simply adding a language specific dialect classifier, provided the language identification model supports that particular language. 
    
\end{enumerate}

%============================== 2 way using 3 way ================================%

% Hyperparameters

\section{Conclusion}
% Aditya & Ankit
In this paper we propose a two-stage classification pipeline for dialect identification for multilingual corpora. We conduct thorough ablations on this setup and provide valuable insights. We foresee multiple future directions for this work. The first is to expand this work to many languages and dialects. Secondly, it is a worthwhile research direction to distill this multi-model setup into a single model with multiple prediction heads. 

\section*{Limitations}
The obvious limitation of this system is the excessive memory consumption due to the usage of language specific models. For low resource languages this system is difficult to train and scale. We hope that these problems will be addressed by researchers in future works.

\bibliography{anthology, custom}
\bibliographystyle{acl_natbib}

\end{document}